\documentclass[11pt]{article}

\usepackage[margin=1in]{geometry}
\usepackage{cite}
\usepackage{amsmath,amssymb}
\usepackage{graphicx}
\usepackage{booktabs}
\usepackage{multirow}
\usepackage{array}
\usepackage{url}
\usepackage{xcolor}
\usepackage{caption}
\usepackage{tikz}
\usepackage{longtable}
\usepackage{hyperref}
\usepackage{float}
\usepackage{longtable}
\usepackage{graphicx}
\usepackage{booktabs}
\usetikzlibrary{arrows.meta, positioning, fit, backgrounds}

\usepackage{xurl}

\usepackage{xltabular}
\usepackage{array}

\newcolumntype{L}{>{\raggedright\arraybackslash}X}

\graphicspath{{paper_assests/figures/}{paper_assests/tables/}}

\title{HypothesisMed: Inference-Time Answer Fusion and Structured Hypothesis-Space Reporting for Biomedical Question Answering}

\author{
Md Motaleb Hossen Manik\\
Department of Computer Science\\
Rensselaer Polytechnic Institute\\
Troy, NY, USA\\
\texttt{manikm@rpi.edu}
\and
Ge Wang\\
Department of Biomedical Engineering\\
Rensselaer Polytechnic Institute\\
Troy, NY, USA\\
\texttt{wangg6@rpi.edu}
}

\date{}

\begin{document}

\maketitle

\begin{abstract}
Biomedical question answering with large language models is commonly evaluated using answer accuracy, but answer accuracy alone does not indicate whether a model can produce parseable outputs, follow structured reliability instructions, recognize weak answer spaces, or avoid confident incorrect commitments. This paper presents HypothesisMed, an inference-time reliability pipeline for biomedical multiple-choice question answering. The pipeline combines direct prompting, chain-of-thought prompting, structured HypothesisMed-v3 prompting, and majority-based answer fusion. The final answer is selected by answer fusion, while HypothesisMed-v3 supplies structured SPACE labels and confidence information. SPACE labels describe whether the answer space is \textit{VALID}, \textit{INCOMPLETE}, or \textit{CONTRADICTED}. We first evaluate four open-weight models, Qwen2.5-7B-Instruct, Phi-4-mini-instruct, DeepSeek-R1-Distill-Qwen-32B, and BioMistral-7B, across MedQA, MedMCQA, and PubMedQA using 1,000 examples per dataset. In this primary four-model evaluation, the proposed answer-fusion plus HypothesisMed-v3 SPACE pipeline improves weighted answer accuracy over each model's best direct or chain-of-thought baseline while increasing parse coverage and structured SPACE coverage. We then conduct a scaled evaluation on Qwen2.5-7B-Instruct and Phi-4-mini-instruct using 10,183 examples per model. In the scaled setting, fusion substantially improves Phi-4-mini accuracy, from 0.4296 to 0.5192 over its best direct or chain-of-thought baseline, while Qwen2.5-7B chain-of-thought prompting remains slightly higher in answer accuracy than fusion. However, Qwen2.5-7B fusion provides complete parse coverage, complete SPACE coverage, and substantially lower false commitment. An expanded 12,000-example SPACE stress test further shows that answer-space diagnosis remains difficult: Qwen2.5-7B reaches 0.3074 SPACE accuracy and Phi-4-mini reaches 0.4168. These results show that answer accuracy, parseability, structured reliability reporting, calibration behavior, and false-commitment behavior are separable capabilities. The main contribution is not a universal state-of-the-art claim, but a reproducible inference-time framework for evaluating biomedical question answering models as auditable workflow components under structured reliability constraints.
\end{abstract}


\section{Introduction}

Large language models (LLMs) are increasingly evaluated for biomedical and clinical question answering, medical education support, and biomedical literature reasoning. Recent medical LLM studies have moved beyond general natural language processing benchmarks and have evaluated models on medical examination questions, biomedical research questions, and consumer health questions \cite{singhal2023clinicalknowledge,singhal2025expertmedicalqa}. In these settings, the usual benchmark objective is to select the correct answer. However, correctness alone is not sufficient for reliability-sensitive biomedical workflows. A model may select the right option but produce an output that cannot be parsed automatically. It may also confidently select an answer even when the available options are incomplete, duplicated, ambiguous, or internally inconsistent.

Biomedical multiple-choice question answering is a useful setting for studying these issues because it requires both domain knowledge and constrained reasoning over a defined answer space. Benchmarks such as MedQA, MedMCQA, and PubMedQA are widely used to evaluate biomedical question answering systems \cite{jin2021medqa,pal2022medmcqa,jin2019pubmedqa}. MedQA uses professional medical board examination questions, MedMCQA contains large-scale medical entrance examination questions, and PubMedQA focuses on biomedical research questions derived from PubMed abstracts \cite{jin2021medqa,pal2022medmcqa,jin2019pubmedqa}. However, many benchmark evaluations primarily report final answer accuracy. Accuracy is necessary, but it does not fully characterize whether a model can follow structured output instructions, provide machine-parseable responses, expose uncertainty, or avoid high-confidence incorrect commitments.

This gap is important because recent LLM evaluation work argues that model assessment should include multiple dimensions beyond accuracy, including calibration, robustness, transparency, and other reliability-oriented properties \cite{liang2023helm}. In biomedical applications, these concerns are amplified because fluent but incorrect model outputs can appear plausible and may be difficult to detect without additional reliability checks. Prior work on hallucination and LLM reliability has similarly emphasized that plausible but unsupported outputs remain a major obstacle to safe deployment \cite{huang2025hallucination}. These concerns motivate evaluation protocols that measure not only whether an answer is correct, but also whether the answer is parseable, auditable, and accompanied by structured reliability information.

This paper addresses these questions through HypothesisMed, an inference-time pipeline for biomedical hypothesis-space reporting. In this paper, SPACE refers to the structured answer-space status field used by HypothesisMed-v3; it is a label name rather than a model name or performance metric. The core idea is to separate answer selection from reliability reporting. Direct prompting and chain-of-thought prompting provide answer candidates. HypothesisMed-v3 provides structured SPACE metadata, where SPACE describes whether the answer space is \textit{VALID}, \textit{INCOMPLETE}, or \textit{CONTRADICTED}. The proposed method then uses answer fusion for final answer selection while retaining HypothesisMed-v3 outputs for reliability analysis. This design is related to prior inference-time reasoning strategies such as chain-of-thought prompting and self-consistency, but it targets structured biomedical reliability reporting rather than reasoning accuracy alone \cite{wei2022cot,wang2023selfconsistency}.

This study asks four research questions:

\begin{enumerate}
    \item In a primary four-model evaluation, does answer fusion plus HypothesisMed-v3 SPACE reporting improve answer accuracy over direct and chain-of-thought baselines?
    \item In a larger scaled evaluation, does the proposed pipeline preserve or improve answer accuracy while increasing parseability and structured reliability observability?
    \item Can SPACE labels detect controlled answer-space defects, including incomplete and contradicted option sets?
    \item Do open-weight biomedical and general-purpose LLMs show separable behavior across answer accuracy, parseability, SPACE-label coverage, calibration behavior, and false-commitment behavior?
\end{enumerate}

The main contributions are:

\begin{enumerate}
    \item We introduce an inference-time reliability pipeline that combines answer fusion with structured hypothesis-space reporting for biomedical question answering.
    \item We define and evaluate SPACE-label behavior for biomedical multiple-choice tasks using \textit{VALID}, \textit{INCOMPLETE}, and \textit{CONTRADICTED} labels.
    \item We evaluate four open-weight models across three biomedical question answering datasets using direct prompting, chain-of-thought prompting, HypothesisMed-v3 prompting, and the proposed fusion method.
    \item We conduct a scaled two-model evaluation with 10,183 examples per model, showing that fusion substantially improves Phi-4-mini accuracy while Qwen2.5-7B fusion remains competitive in answer accuracy and provides higher parse coverage, higher SPACE coverage, and lower false commitment than direct or chain-of-thought baselines.
    \item We construct an expanded 12,000-example SPACE stress test per model, showing that answer-space diagnosis is measurable but remains difficult for prompt-only structured inference.
    \item We provide a reliability-oriented analysis showing that answer accuracy, parse coverage, SPACE coverage, SPACE accuracy, calibration behavior, and false commitment should be treated as distinct evaluation axes.
\end{enumerate}

The rest of the paper is organized as follows. Section~\ref{sec:related_work} reviews related work on biomedical question answering benchmarks, prompting and inference-time ensembling, medical LLMs, structured output, calibration, and reliability evaluation. Section~\ref{sec:methodology} presents the HypothesisMed methodology, including the workflow, datasets, models, prompting methods, SPACE-label definitions, expanded SPACE stress test, and evaluation metrics. Section~\ref{sec:results} reports the experimental results, including the primary four-model evaluation, scaled two-model evaluation, per-dataset analysis, SPACE stress-test results, structured reliability behavior, confidence and calibration analysis, and positioning against published results. Section~\ref{sec:discussion} discusses the implications of the findings, limitations of the current study, and directions for future work. Section~\ref{sec:conclusion} concludes the paper by summarizing the main contributions and emphasizing the need to evaluate biomedical LLMs as auditable workflow components rather than answer generators alone.

\section{Related Work}
\label{sec:related_work}

\subsection{Biomedical Question Answering Benchmarks}

Biomedical question answering benchmarks provide standardized tasks for evaluating medical knowledge, clinical reasoning, and biomedical evidence interpretation. MedQA introduced a multilingual medical question answering benchmark based on professional medical board examinations \cite{jin2021medqa}. MedMCQA provides a large-scale multiple-choice dataset derived from medical entrance examinations and covers a broad range of healthcare topics and medical subjects \cite{pal2022medmcqa}. PubMedQA evaluates biomedical research question answering over PubMed abstracts and requires yes, no, or maybe answers grounded in the corresponding abstract context \cite{jin2019pubmedqa}. These datasets are frequently used in medical LLM evaluation because they cover complementary aspects of biomedical reasoning: examination-style medical knowledge, multi-subject medical question answering, and biomedical literature interpretation.

Several medical LLM studies have combined these datasets into broader evaluation suites. For example, MultiMedQA includes existing medical question answering datasets and additional consumer medical questions to evaluate medical knowledge and answer quality along multiple axes \cite{singhal2023clinicalknowledge}. Med-PaLM 2 further showed that medical-domain adaptation, prompting strategies, and ensemble-style refinement can substantially improve medical question answering performance \cite{singhal2025expertmedicalqa}. These studies demonstrate the value of biomedical QA benchmarks, but they also show that medical LLM evaluation increasingly requires more than a single accuracy score. The present work builds on these benchmark traditions while focusing on parseability, structured answer-space reporting, and false-commitment behavior.

\subsection{Prompting, Reasoning, and Inference-Time Ensembling}

Prompting strategies are central to LLM-based question answering. Direct prompting asks the model to produce an answer with minimal intermediate reasoning. Chain-of-thought prompting elicits intermediate reasoning steps before the final answer and has been shown to improve performance on multi-step reasoning tasks \cite{wei2022cot}. Self-consistency extends chain-of-thought prompting by sampling multiple reasoning paths and selecting the most consistent answer, improving reasoning performance across several benchmark settings \cite{wang2023selfconsistency}. These methods show that inference-time procedures can change model behavior without modifying model weights.

Biomedical question answering can benefit from inference-time reasoning and ensembling, but medical workflows also require reliable formatting and interpretable output behavior. A chain-of-thought response may improve final answer accuracy while producing long or inconsistent text that is harder to parse automatically. Similarly, majority-based answer fusion may improve answer selection but does not by itself indicate whether the model recognized a weak, incomplete, or contradicted answer space. HypothesisMed builds on the inference-time prompting tradition but separates final answer selection from structured reliability reporting.

\subsection{Medical LLMs and Domain-Specialized Models}

Medical LLM research has explored both general-purpose instruction-tuned models and biomedical-domain models. Med-PaLM and Med-PaLM 2 showed that large general-purpose models can be adapted and prompted for medical question answering, with evaluation spanning professional medical exams, biomedical research questions, and consumer health questions \cite{singhal2023clinicalknowledge,singhal2025expertmedicalqa}. Open-weight general-purpose model families such as Qwen2.5 and Phi-4 provide instruction-following baselines that can be evaluated locally and reproducibly \cite{yang2024qwen25,abdin2024phi4}. Reasoning-oriented models such as DeepSeek-R1 introduce another model class in which reinforcement learning is used to strengthen reasoning behavior \cite{deepseek2025r1}. Biomedical-domain models such as BioMistral are adapted from general language models using biomedical corpora and evaluated on medical QA tasks \cite{labrak2024biomistral}.

These model families may differ not only in answer accuracy but also in instruction following, output format compliance, and structured reasoning behavior. This distinction is important for the present study because a model can have biomedical knowledge but still fail to produce parseable structured outputs. Conversely, a general instruction-tuned model may behave more reliably in an automated workflow even if it is not explicitly biomedical-domain specialized. Therefore, this paper evaluates models as workflow components, not only as answer generators.

\subsection{Structured Output, Calibration, and Reliability Evaluation}

Modern LLM evaluation increasingly emphasizes reliability dimensions beyond accuracy. The Holistic Evaluation of Language Models framework argues for multi-metric evaluation, including accuracy, calibration, robustness, fairness, bias, toxicity, and efficiency \cite{liang2023helm}. This perspective is especially relevant in biomedical settings because a correct final answer does not guarantee that the response is parseable, calibrated, or safe to integrate into downstream workflows.

Structured output reliability is also important for automated biomedical evaluation. When a model is expected to produce a machine-readable answer, an unparseable output can become an operational failure even if the text contains useful information. More broadly, hallucination research has shown that LLMs can generate fluent but factually unsupported or inconsistent outputs, which motivates explicit reliability checks and uncertainty-aware evaluation \cite{huang2025hallucination}. In medicine, uncertainty communication is especially important because incorrect confidence can mislead users and obscure model limitations \cite{savage2024uncertainty}. HypothesisMed contributes to this reliability-oriented direction by evaluating answer accuracy together with parse coverage, SPACE-label coverage, SPACE accuracy, calibration behavior, and false-commitment behavior.

\subsection{Positioning of HypothesisMed}

HypothesisMed is not intended to replace medical-domain training, expert clinical validation, or full safety evaluation. Instead, it provides an inference-time reliability layer for biomedical multiple-choice question answering. It builds on prior biomedical QA benchmarks, medical LLM evaluation, inference-time prompting, and reliability-oriented evaluation, but focuses on a narrower structured-output question: whether biomedical QA systems can provide parseable answers together with answer-space status and high-confidence error diagnostics. Table~\ref{tab:conceptual_comparison} summarizes this positioning.

\begin{table*}[t]
\centering
\small
\caption{Conceptual positioning of HypothesisMed relative to cited benchmark, prompting, medical LLM, and reliability-evaluation work.}
\label{tab:conceptual_comparison}
\begin{tabular}{p{3.0cm}p{4.3cm}p{3.1cm}p{4cm}}
\toprule
Literature category & Main role in this paper & Answer-space status & Confidence/error behavior \\ 
\midrule
Biomedical QA benchmarks \cite{jin2021medqa,pal2022medmcqa,jin2019pubmedqa}
& Provide standardized biomedical QA tasks
& Not part of original benchmark task
& Not part of original benchmark task \\

Medical LLM evaluation \cite{singhal2023clinicalknowledge,singhal2025expertmedicalqa}
& Motivates medical QA evaluation and multi-axis answer-quality assessment
& Not reported as SPACE labels
& Evaluates safety/quality dimensions, but not this false-commitment metric \\

Prompting and self-consistency \cite{wei2022cot,wang2023selfconsistency}
& Motivates inference-time answer selection methods
& Not explicitly targeted
& Not explicitly targeted \\

Reliability and uncertainty evaluation \cite{liang2023helm,huang2025hallucination,savage2024uncertainty}
& Motivates evaluation beyond accuracy, including calibration, hallucination, and uncertainty
& Related, but not specific to biomedical MCQ answer-space validity
& Closely related; operationalized here as false commitment \\

HypothesisMed
& Adds structured reliability reporting to biomedical QA
& VALID, INCOMPLETE, CONTRADICTED
& Measures high-confidence wrong answers through false commitment \\
\bottomrule
\end{tabular}
\end{table*}

\section{Methodology}
\label{sec:methodology}
\subsection{Workflow Overview}

\begin{figure}
    \centering
    \includegraphics[width=\linewidth]{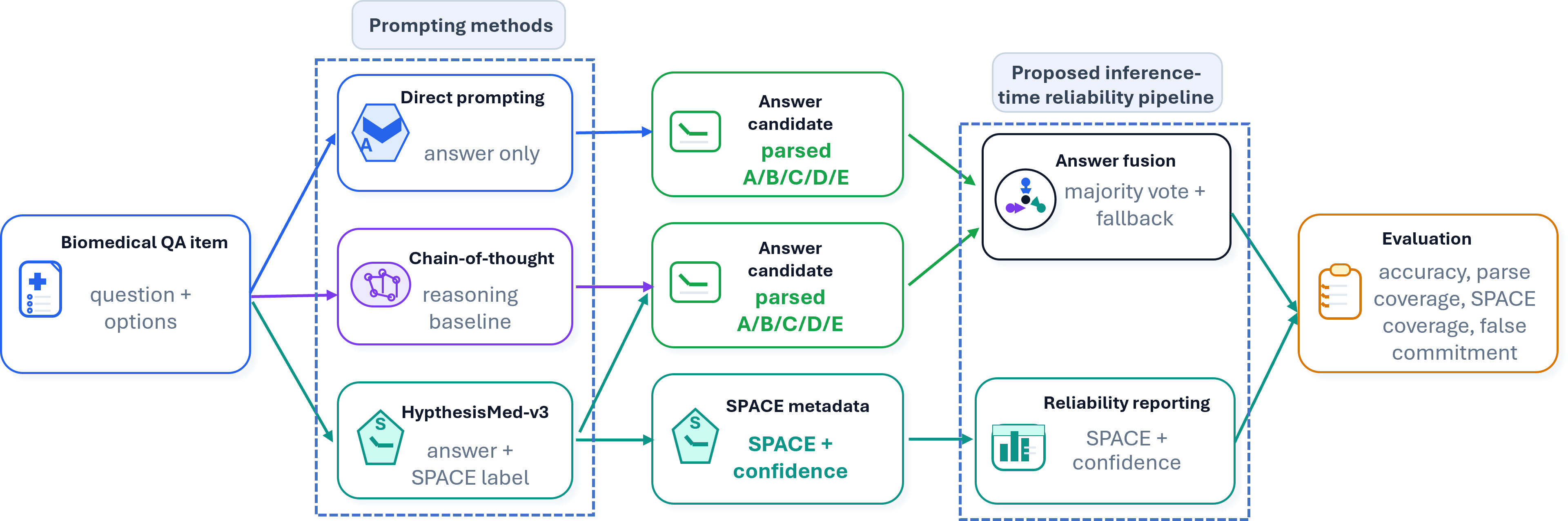}
    \caption{Overview of the HypothesisMed inference-time reliability pipeline. Direct prompting, chain-of-thought prompting, and HypothesisMed-v3 prompting are applied to each biomedical question. The final answer is selected using answer fusion over parseable answer candidates, while HypothesisMed-v3 also supplies structured SPACE labels and confidence values for reliability analysis.}
    \label{fig:workflow}
\end{figure}

Figure~\ref{fig:workflow} summarizes the HypothesisMed pipeline. Each biomedical question is evaluated using direct prompting, chain-of-thought prompting, and HypothesisMed-v3 prompting. The proposed method fuses answer candidates using majority voting and attaches HypothesisMed-v3 SPACE labels and confidence metadata for reliability analysis.

\subsection{Datasets and Evaluation Splits}

We evaluate HypothesisMed on three biomedical question answering benchmarks: MedQA, MedMCQA, and PubMedQA. MedQA contains medical examination-style multiple-choice questions, MedMCQA contains large-scale medical multiple-choice questions spanning multiple medical subjects, and PubMedQA contains biomedical research questions derived from PubMed abstracts \cite{jin2021medqa,pal2022medmcqa,jin2019pubmedqa}.

The primary four-model experiment uses fixed 1,000-example subsets from each dataset, yielding 3,000 examples per model. These deterministic subsets are named \texttt{medqa\_original1000}, \texttt{medmcqa\_original1000}, and \texttt{pubmedqa\_original1000}. Each example contains a question, candidate answer options, and the benchmark gold answer. The reported aggregate metrics are weighted by the number of examples, which is equal across datasets in the primary experiment.

To address the limited size of the 1,000-example subsets, we also conduct a scaled evaluation on the two strongest structured-output models, Qwen2.5-7B-Instruct and Phi-4-mini-instruct. This scaled evaluation uses 5,000 MedQA examples, 4,183 MedMCQA examples, and 1,000 PubMedQA examples, for a total of 10,183 examples per model. The scaled evaluation is used to test whether the main conclusions persist beyond the original fixed 1,000-example subsets.

Because the original benchmark examples contain unmodified answer options, the gold SPACE label for these main benchmark evaluations is \textit{VALID}. Therefore, SPACE accuracy on the original benchmark examples measures whether the model correctly recognizes standard benchmark answer spaces as valid when a SPACE label is extracted. This design tests whether structured reliability prompting introduces spurious claims of incompleteness or contradiction on standard biomedical QA inputs.

\subsection{Expanded SPACE Stress Test}

The original benchmark evaluation alone cannot fully validate whether SPACE labels detect genuinely defective answer spaces, because all original benchmark items are treated as \textit{VALID}. We therefore construct an expanded controlled SPACE stress test from MedQA, MedMCQA, and PubMedQA. For each selected original item, we generate four answer-space variants: the original \textit{VALID} item, an \textit{INCOMPLETE} item created by removing the correct answer option, a \textit{CONTRADICTED} item created by duplicating the correct answer under another option letter, and a second \textit{CONTRADICTED} item created by making the answer space non-unique through another duplicated correct option. The stress test uses 1,000 base examples from each dataset, yielding 12,000 stress-test examples per model. This stress test is balanced by stress type rather than by SPACE label, because two contradiction-style perturbations are included for each base item.

\subsection{Models}

We evaluate four open-weight LLMs in the primary experiment: Qwen2.5-7B-Instruct, Phi-4-mini-instruct, DeepSeek-R1-Distill-Qwen-32B, and BioMistral-7B. Qwen2.5-7B-Instruct and Phi-4-mini-instruct are general instruction-tuned models, DeepSeek-R1-Distill-Qwen-32B is a reasoning-oriented distilled model, and BioMistral-7B is a biomedical-domain model \cite{yang2024qwen25,abdin2024phi4,deepseek2025r1,labrak2024biomistral}. The scaled evaluation and expanded SPACE stress test focus on Qwen2.5-7B-Instruct and Phi-4-mini-instruct because they show the strongest structured-output behavior in the primary experiment, based on their higher parse and SPACE coverage in the primary evaluation.

All models were evaluated locally using vLLM, a high-throughput LLM inference engine based on PagedAttention \cite{kwon2023vllm}. Each model was evaluated under direct prompting, chain-of-thought prompting, HypothesisMed-v3 structured prompting, and the proposed answer-fusion plus Hypothe-sisMed-v3 SPACE pipeline. The same dataset examples were used across methods for each model, ensuring paired comparisons at the example level.

All experiments were run locally using vLLM with deterministic decoding. The main inference configuration used bfloat16 precision, a maximum context length of 4096 tokens, temperature 0.0, top-$p$ 1.0, and a maximum generation length of 512 tokens in the final cached-model evaluation scripts. Dataset sampling used seed 42. The experiments were run on an axis2 compute node with eight NVIDIA H100 80GB HBM3 GPUs, dual Intel Xeon Platinum 8468 processors, and 1.0 TiB system memory. These settings are summarized in Table~\ref{tab:settings}.

\begin{table}[t]
\centering
\caption{Inference, evaluation, and compute settings used in the experiments.}
\label{tab:settings}
\begin{tabular}{lp{10cm}}
\toprule
Setting & Value \\
\midrule
Inference engine & vLLM 0.19.1 \\
Precision & bfloat16 \\
Maximum context length & 4096 tokens \\
Maximum generation length & 512 tokens \\
Temperature & 0.0 \\
Top-$p$ & 1.0 \\
Dataset sampling seed & 42 \\
Primary datasets & MedQA, MedMCQA, PubMedQA \\
Primary examples per dataset & 1,000 \\
Scaled examples & 5,000 MedQA; 4,183 MedMCQA; 1,000 PubMedQA \\
Expanded SPACE stress test & 12,000 examples per model \\
Primary models & Qwen2.5-7B, Phi-4-mini, DeepSeek-R1-32B, BioMistral-7B \\
Scaled/stress-test models & Qwen2.5-7B, Phi-4-mini \\
Prompting methods & Direct, CoT, HypMed-v3, Proposed fusion \\
Fusion rule & Majority vote with parseable fallback \\
Gold SPACE label & VALID for original benchmark items; perturbation-defined for stress test \\
False-commitment threshold & Confidence $\geq 0.5$ \\
Compute node & axis2 \\
GPU configuration & 8 $\times$ NVIDIA H100 80GB HBM3 \\
CPU & Dual Intel Xeon Platinum 8468 \\
System memory & 1.0 TiB \\
Operating system & Ubuntu 22.04.5 LTS \\
CUDA / driver & CUDA 12.4 / NVIDIA driver 550.144.03 \\
\bottomrule
\end{tabular}
\vspace{1mm}

\footnotesize{
\textit{Notes.} CoT denotes chain-of-thought prompting. HypMed-v3 denotes structured HypothesisMed-v3 prompting. Temperature, top-$p$, context length, precision, and generation length were extracted from the experiment scripts and vLLM runner configuration.}
\end{table}

The final cached-model evaluation scripts used a maximum generation length of 512 tokens; earlier smoke-test utilities used shorter defaults and are not used for the reported final results.

\subsection{Prompting Methods}

We evaluate four inference-time method families.

\textbf{Direct prompting} asks the model to select the best answer from the provided options with minimal reasoning. The parser extracts the final selected option from the model output.

\textbf{Chain-of-thought prompting} asks the model to reason through the question before selecting the final answer. This baseline tests whether explicit reasoning improves final answer selection, while also allowing us to measure whether longer reasoning outputs reduce parseability.

We use SPACE to denote the structured answer-space status field returned by HypothesisMed-v3. SPACE is treated as a label name for answer-space validity rather than as an independent model or calibrated uncertainty score. The SPACE label takes one of three values:

\begin{itemize}
    \item \textit{VALID}: the option set contains one medically supported best answer.
    \item \textit{INCOMPLETE}: the correct answer appears missing or insufficiently represented.
    \item \textit{CONTRADICTED}: the available options are internally inconsistent, duplicated in a way that prevents a unique answer, or medically contradictory.
\end{itemize}

\textbf{Proposed answer fusion plus HypothesisMed-v3 SPACE} combines answer candidates from direct prompting, chain-of-thought prompting, and HypothesisMed-v3 prompting. The final answer is selected using majority voting over parseable answer candidates. When no strict majority exists, the pipeline uses a deterministic parseable fallback order: direct prompting, chain-of-thought prompting, and then HypothesisMed-v3. HypothesisMed-v3 supplies the structured SPACE label and confidence metadata used for reliability analysis.

\subsection{Evaluation Metrics}

We report five main evaluation metrics.

\textbf{Answer accuracy} is the fraction of examples for which the predicted answer matches the benchmark gold answer.

\textbf{Parse coverage} is the fraction of examples for which the evaluation parser extracts a valid answer option from the model output. Low parse coverage indicates that a model may have generated text that is difficult to use in an automated workflow, even if the text contains useful reasoning.

\textbf{SPACE coverage} is the fraction of examples for which a valid SPACE label is extracted. In the main benchmark experiments, all original unmodified questions have gold SPACE label \textit{VALID}. In the expanded SPACE stress test, gold SPACE labels are assigned by the controlled perturbation type.

\textbf{SPACE accuracy} is the fraction of extracted SPACE labels that match the gold SPACE label. On original benchmark examples, this measures whether the model avoids incorrectly labeling standard benchmark questions as \textit{INCOMPLETE} or \textit{CONTRADICTED}. On the expanded SPACE stress test, it measures whether the model identifies controlled answer-space defects.

\textbf{False commitment} measures high-confidence wrong-answer behavior. For structured Hypothe-sisMed-v3 outputs, an incorrect response is counted as a false commitment when the extracted confidence score is at least 0.5. For direct and chain-of-thought baselines, a parseable final answer without an extracted uncertainty signal is treated as a committed answer. Lower false commitment is better. This metric is intended to measure whether a method makes wrong answers while still appearing decisive.

We also report paired exact McNemar tests for selected answer-accuracy comparisons, fallback-order sensitivity for fusion, calibration summaries using ten-bin expected calibration error, and structured-output failure statistics in the appendix.

\section{Results}
\label{sec:results}

\subsection{Primary Four-Model Results on 1,000-Example Subsets}

Table~\ref{tab:aggregate} reports aggregate results across MedQA, MedMCQA, and PubMedQA. Each model is evaluated on 3,000 examples, with 1,000 examples from each dataset. In this primary four-model setting, the proposed answer-fusion plus HypothesisMed-v3 SPACE method improves weighted answer accuracy over each model's best direct or chain-of-thought baseline, while also increasing parse coverage and SPACE-label coverage.

\begin{table*}[t]
\centering
\small
\caption{Aggregate four-model results across MedQA, MedMCQA, and PubMedQA. Each row aggregates 3,000 examples from three datasets. Higher accuracy, parse coverage, and SPACE coverage are better. Lower false commitment is better.}
\label{tab:aggregate}

\begin{tabular}{llrrrr}
\toprule
Model & Method & Accuracy & Parse cov. & SPACE cov. & False commit. \\
\midrule
Qwen2.5-7B & Proposed & 0.6367 & 1.0000 & 0.9997 & 0.1026 \\
Qwen2.5-7B & CoT & 0.6003 & 0.9440 & 0.0020 & 0.8478 \\
Qwen2.5-7B & Direct & 0.5650 & 0.9147 & 0.0007 & 0.9255 \\
Qwen2.5-7B & HypMed-v3 & 0.5623 & 0.9860 & 0.9997 & 0.0973 \\
\midrule
Phi-4-mini & Proposed & 0.5767 & 0.9913 & 0.9920 & 0.4054 \\
Phi-4-mini & CoT & 0.4817 & 0.7923 & 0.0000 & 0.9838 \\
Phi-4-mini & Direct & 0.3020 & 0.5403 & 0.0000 & 0.8329 \\
Phi-4-mini & HypMed-v3 & 0.4943 & 0.9310 & 0.9920 & 0.4341 \\
\midrule
DeepSeek-R1-32B & Proposed & 0.3877 & 0.7213 & 0.4430 & 0.0213 \\
DeepSeek-R1-32B & CoT & 0.0710 & 0.2370 & 0.0010 & 0.0000 \\
DeepSeek-R1-32B & Direct & 0.1690 & 0.3557 & 0.0010 & 0.2986 \\
DeepSeek-R1-32B & HypMed-v3 & 0.3413 & 0.5793 & 0.4430 & 0.0266 \\
\midrule
BioMistral-7B & Proposed & 0.1307 & 0.3330 & 0.0040 & 0.0000 \\
BioMistral-7B & CoT & 0.1247 & 0.3203 & 0.0000 & 0.8287 \\
BioMistral-7B & Direct & 0.0080 & 0.0147 & 0.0000 & 0.1310 \\
BioMistral-7B & HypMed-v3 & 0.0007 & 0.0040 & 0.0040 & 0.0000 \\
\bottomrule
\end{tabular}

\vspace{2mm}

{\footnotesize
\textit{Notes.} Each model is evaluated on 3,000 examples: 1,000 each from MedQA, MedMCQA, and PubMedQA. Proposed denotes answer fusion plus HypothesisMed-v3 SPACE reporting. CoT denotes chain-of-thought prompting. HypMed-v3 denotes structured HypothesisMed-v3 prompting without answer fusion. Parse cov. denotes parse coverage, SPACE cov. denotes SPACE-label coverage, and false commit. denotes false commitment.
}
\end{table*}

Qwen2.5-7B achieves the strongest aggregate performance under the proposed method, with 0.6367 weighted answer accuracy, 1.0000 parse coverage, and 0.9997 SPACE coverage. Phi-4-mini also benefits substantially from the proposed method, reaching 0.5767 weighted answer accuracy and 0.9920 SPACE coverage. HypothesisMed-v3 alone provides high SPACE coverage for Qwen2.5-7B and Phi-4-mini, but it does not always maximize answer accuracy. This supports the design choice of separating answer selection from structured reliability reporting.

\subsection{Accuracy Gains, Reliability Deltas, and Fusion Ablation}

Table~\ref{tab:deltas} compares the proposed method with each model's best direct or chain-of-thought baseline in the primary four-model experiment.

\begin{table*}[t]
\centering
\small
\caption{Accuracy and reliability deltas of the proposed method compared with each model's best direct or chain-of-thought baseline.}
\label{tab:deltas}

\begin{tabular}{lllll}
\toprule
Metric & Qwen2.5-7B & Phi-4-mini & DeepSeek-R1-32B & BioMistral-7B \\
\midrule
Best baseline & CoT & CoT & Direct & CoT \\
Proposed accuracy & 0.6367 & 0.5767 & 0.3877 & 0.1307 \\
Baseline accuracy & 0.6003 & 0.4817 & 0.1690 & 0.1247 \\
Absolute accuracy gain & 0.0364 & 0.0950 & 0.2187 & 0.0060 \\
Relative accuracy gain (\%) & 6.06 & 19.72 & 129.41 & 4.81 \\
\midrule
Proposed parse coverage & 1.0000 & 0.9913 & 0.7213 & 0.3330 \\
Baseline parse coverage & 0.9440 & 0.7923 & 0.3557 & 0.3203 \\
Proposed SPACE coverage & 0.9997 & 0.9920 & 0.4430 & 0.0040 \\
Baseline SPACE coverage & 0.0020 & 0.0000 & 0.0010 & 0.0000 \\
\midrule
Proposed false commitment & 0.1026 & 0.4054 & 0.0213 & 0.0000 \\
Baseline false commitment & 0.8478 & 0.9838 & 0.2986 & 0.8287 \\
False commitment reduction & 0.7452 & 0.5784 & 0.2773 & 0.8287 \\
\bottomrule
\end{tabular}

\vspace{2mm}

\footnotesize{
\textit{Notes.} Proposed denotes answer fusion plus HypothesisMed-v3 SPACE reporting. CoT denotes chain-of-thought prompting. The best baseline is selected from direct prompting and chain-of-thought prompting for each model. SPACE coverage denotes the fraction of examples with an extracted SPACE label. False commitment is lower when the model makes fewer high-confidence wrong commitments.}
\end{table*}

For Qwen2.5-7B, the proposed method improves weighted answer accuracy from 0.6003 to 0.6367 relative to the best baseline. For Phi-4-mini, the proposed method improves accuracy from 0.4817 to 0.5767. For DeepSeek-R1-32B, the proposed method improves accuracy from 0.1690 to 0.3877, but this result should be interpreted cautiously because the direct and chain-of-thought baselines have low parse coverage. We therefore treat DeepSeek-R1-32B primarily as a formatting-stress-test case: the result shows improved automated usability under difficult output-format behavior, not a clean biomedical reasoning gain of the same relative magnitude. BioMistral-7B shows the weakest structured-output performance, with low parse coverage and near-zero SPACE coverage. This should not be interpreted as a general claim that BioMistral lacks biomedical knowledge; rather, it indicates a structured-output compliance failure in this evaluation pipeline.

Table~\ref{tab:fusion_ablation} separates the contribution of HypothesisMed-v3 from the contribution of answer fusion. HypothesisMed-v3 alone provides structured SPACE reporting, while the proposed method combines answer fusion with HypothesisMed-v3 SPACE metadata. For example, Qwen2.5-7B obtains 0.5623 accuracy with HypothesisMed-v3 alone, below its CoT baseline, but reaches 0.6367 with the proposed fusion pipeline. This supports the design choice that HypothesisMed-v3 is not intended to be the strongest standalone answer selector. Its role is to provide structured answer-space metadata that can be combined with answer fusion.

\begin{table*}[t]
\centering
\small
\caption{Fusion ablation across three datasets. HypMed-v3 alone provides structured SPACE reporting, while the proposed method combines answer fusion with HypMed-v3 SPACE reporting.}
\label{tab:fusion_ablation}

\begin{tabular}{p{2.8cm}p{2.1cm}p{2.1cm}p{2.1cm}p{2.1cm}p{2.1cm}}
\toprule
Model & Best D/CoT acc. & HypMed-v3 acc. & Proposed acc. & Gain vs. baseline & Gain vs. HypMed-v3 \\
\midrule
Qwen2.5-7B & 0.6003 & 0.5623 & 0.6367 & 0.0364 & 0.0744 \\
Phi-4-mini & 0.4817 & 0.4943 & 0.5767 & 0.0950 & 0.0824 \\
DeepSeek-R1-32B & 0.1690 & 0.3413 & 0.3877 & 0.2187 & 0.0464 \\
BioMistral-7B & 0.1247 & 0.0007 & 0.1307 & 0.0060 & 0.1300 \\
\bottomrule
\end{tabular}

\vspace{1mm}

{\footnotesize
\textit{Notes.} Best D/CoT acc. is the stronger result from direct prompting and chain-of-thought prompting for each model. HypMed-v3 acc. is structured HypothesisMed-v3 prompting without answer fusion. Proposed acc. is answer fusion plus HypothesisMed-v3 SPACE reporting.
}
\end{table*}

Additional paired significance tests and fallback-order sensitivity analyses are provided in the Appendix.

\subsection{Scaled Evaluation on Larger Splits}

\begin{figure}[t]
    \centering
    \includegraphics[width=.7\linewidth]{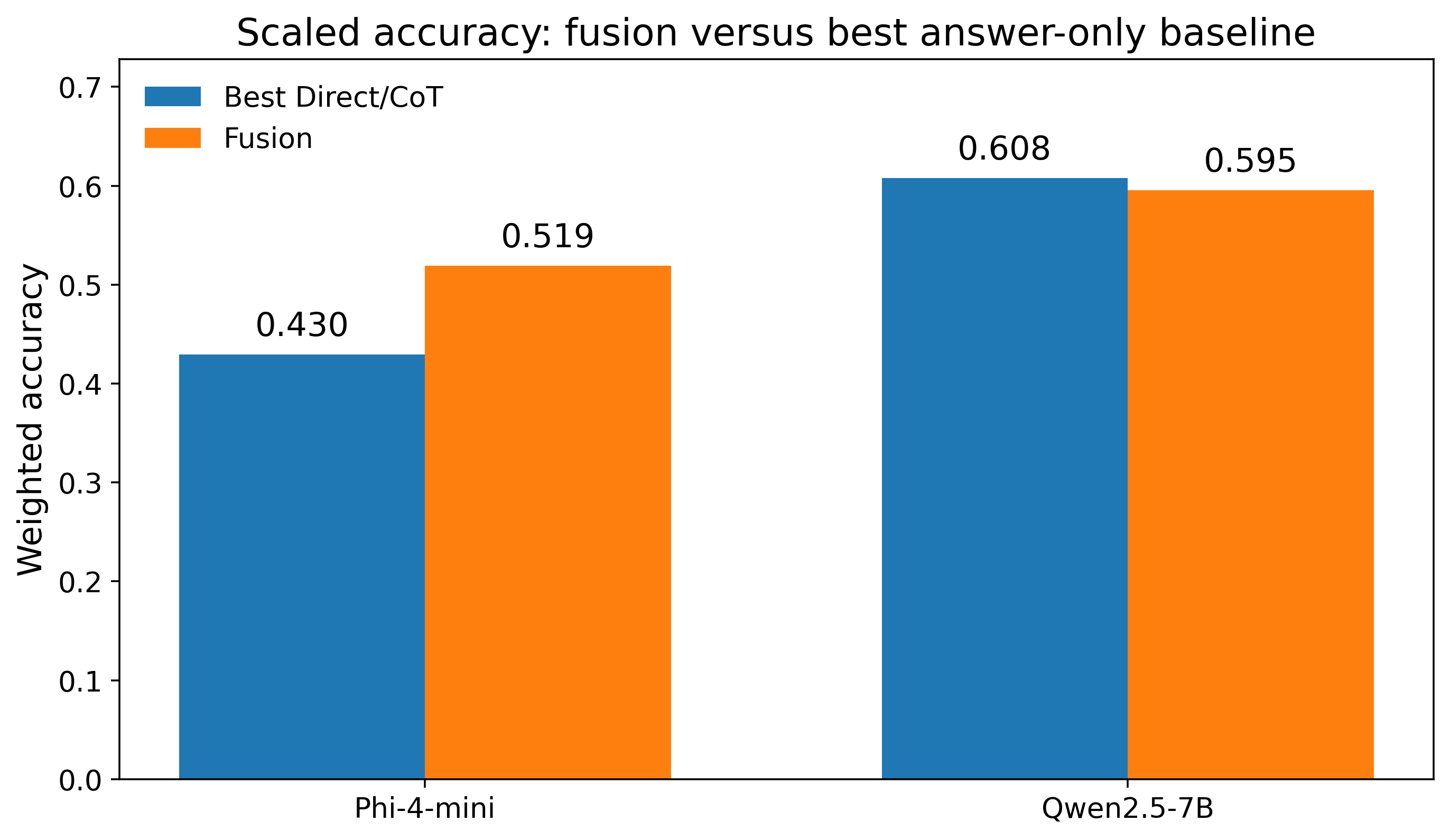}
    \caption{Scaled answer-accuracy comparison between the proposed fusion pipeline and each model's best direct or chain-of-thought baseline.}
    \label{fig:scaled_accuracy}
\end{figure}

To test whether the primary findings persist beyond 1,000-example subsets, we performed a scaled evaluation for Qwen2.5-7B and Phi-4-mini using 5,000 MedQA examples, 4,183 MedMCQA examples, and 1,000 PubMedQA examples. This yields 10,183 examples per model. Table~\ref{tab:scaled_compact} summarizes the scaled comparison between the proposed fusion pipeline and each model's best direct or chain-of-thought baseline, and Figure~\ref{fig:scaled_accuracy} visualizes the corresponding accuracy comparison.

\begin{table*}[t]
\centering
\small
\caption{Scaled evaluation comparing the proposed fusion pipeline with each model's best direct or chain-of-thought baseline across 10,183 examples per model.}
\label{tab:scaled_compact}

\begin{tabular}{lp{1.8cm}p{1.4cm}p{1.4cm}p{1.2cm}p{1.2cm}p{1.4cm}p{1.4cm}}
\toprule
Model & Baseline & Baseline acc. & Fusion acc. & $\Delta$ acc. & Parse cov. & SPACE cov. & False commit. \\\midrule
Phi-4-mini & CoT & 0.4296 & 0.5192 & 0.0896 & 0.9928 & 0.9905 & 0.3271 \\
Qwen2.5-7B & CoT & 0.6079 & 0.5953 & -0.0126 & 1.0000 & 1.0000 & 0.1146 \\
\bottomrule
\end{tabular}

\end{table*}

The scaled results refine the primary conclusion. For Phi-4-mini, the proposed fusion pipeline improves weighted accuracy from 0.4296 for the best direct or chain-of-thought baseline to 0.5192, an absolute gain of 0.0896. It also provides 0.9928 parse coverage and 0.9905 SPACE coverage. For Qwen2.5-7B, chain-of-thought prompting remains slightly higher in answer accuracy than fusion, with 0.6079 weighted accuracy compared with 0.5953 for fusion. However, fusion provides complete parse coverage and complete SPACE coverage, while reducing false commitment from approximately 0.998 for direct or chain-of-thought prompting to 0.1146. Thus, on larger splits, fusion should not be interpreted as uniformly accuracy-dominant. Its stronger and more consistent contribution is workflow completeness, structured observability, and reduced high-confidence wrong commitment.

Paired exact tests on the scaled outputs further clarify this pattern. For Phi-4-mini, fusion significantly outperforms the CoT baseline on MedQA and MedMCQA, with exact $p$-values below $10^{-68}$ in both cases, while PubMedQA shows no significant difference. For Qwen2.5-7B, fusion is significantly below CoT on MedQA, statistically indistinguishable from CoT on MedMCQA and PubMedQA, and still provides stronger structured reliability reporting. The full scaled McNemar test table and fallback-order sensitivity results are reported in the Appendix.

\subsection{Per-Dataset Proposed-Method Results}

Table~\ref{tab:perdataset} reports the proposed method by dataset and model in the primary 1,000-example evaluation. Qwen2.5-7B achieves 0.561 on MedMCQA, 0.596 on MedQA, and 0.753 on PubMedQA. Phi-4-mini achieves 0.489 on MedMCQA, 0.517 on MedQA, and 0.724 on PubMedQA. DeepSeek-R1-32B performs best on PubMedQA among its three dataset results, reaching 0.568. BioMistral-7B remains weak under the proposed structured-output setting.

\begin{table*}[t]
\centering
\small
\caption{Per-dataset results for the proposed answer-fusion plus HypothesisMed-v3 SPACE method in the primary four-model evaluation. Each row uses 1,000 examples. Higher accuracy, parse coverage, SPACE coverage, and SPACE accuracy are better. Lower false commitment is better.}
\label{tab:perdataset}
\resizebox{\textwidth}{!}{%
\begin{tabular}{llrrrrrr}
\toprule
Dataset & Model & Accuracy & 95\% CI & Parse cov. & SPACE cov. & SPACE acc. & False commit. \\
\midrule
MedMCQA & Qwen2.5-7B & 0.561 & [0.530, 0.592] & 1.000 & 1.000 & 0.6350 & 0.0501 \\
MedMCQA & Phi-4-mini & 0.489 & [0.458, 0.520] & 0.997 & 0.994 & 0.5191 & 0.3465 \\
MedMCQA & DeepSeek-R1-32B & 0.340 & [0.311, 0.369] & 0.819 & 0.260 & 0.9923 & 0.0522 \\
MedMCQA & BioMistral-7B & 0.164 & [0.141, 0.187] & 0.476 & 0.011 & 0.0000 & 0.0000 \\
\midrule
MedQA & Qwen2.5-7B & 0.596 & [0.566, 0.626] & 1.000 & 0.999 & 0.4324 & 0.1361 \\
MedQA & Phi-4-mini & 0.517 & [0.486, 0.548] & 0.990 & 0.982 & 0.4216 & 0.2156 \\
MedQA & DeepSeek-R1-32B & 0.255 & [0.228, 0.282] & 0.594 & 0.404 & 0.9950 & 0.0118 \\
MedQA & BioMistral-7B & 0.170 & [0.147, 0.193] & 0.420 & 0.001 & 1.0000 & 0.0000 \\
\midrule
PubMedQA & Qwen2.5-7B & 0.753 & [0.726, 0.780] & 1.000 & 1.000 & 0.9060 & 0.1215 \\
PubMedQA & Phi-4-mini & 0.724 & [0.696, 0.752] & 0.987 & 1.000 & 0.9200 & 0.6540 \\
PubMedQA & DeepSeek-R1-32B & 0.568 & [0.537, 0.599] & 0.751 & 0.665 & 0.9970 & 0.0000 \\
PubMedQA & BioMistral-7B & 0.058 & [0.044, 0.072] & 0.103 & 0.000 & -- & 0.0000 \\
\bottomrule
\end{tabular}
}
\vspace{2mm}

{\footnotesize
\textit{Notes.} Each row uses 1,000 examples. Parse cov. denotes parse coverage, SPACE cov. denotes SPACE-label coverage, SPACE acc. denotes SPACE-label accuracy, and false commit. denotes false commitment. The symbol ``--'' indicates that SPACE accuracy is not defined because no SPACE labels were extracted for that setting.
}
\end{table*}

The SPACE accuracy results show dataset-specific behavior. Qwen2.5-7B and Phi-4-mini obtain high SPACE coverage on MedQA but lower SPACE accuracy than on PubMedQA. Because all original benchmark questions are treated as \textit{VALID}, a low SPACE accuracy means that the model often labeled a standard MedQA answer space as \textit{INCOMPLETE} or \textit{CONTRADICTED}. We interpret this as evidence that MedQA's examination-style questions and answer options are more likely to trigger spurious answer-space concerns under structured prompting. This does not imply lower answer accuracy, since Qwen2.5-7B still achieves 0.596 answer accuracy on MedQA under the proposed method. Instead, it shows that SPACE labeling is a separate reliability behavior from answer selection.

\subsection{Expanded SPACE Stress-Test Results}

\begin{figure}[t]
    \centering
    \includegraphics[width=.7\linewidth]{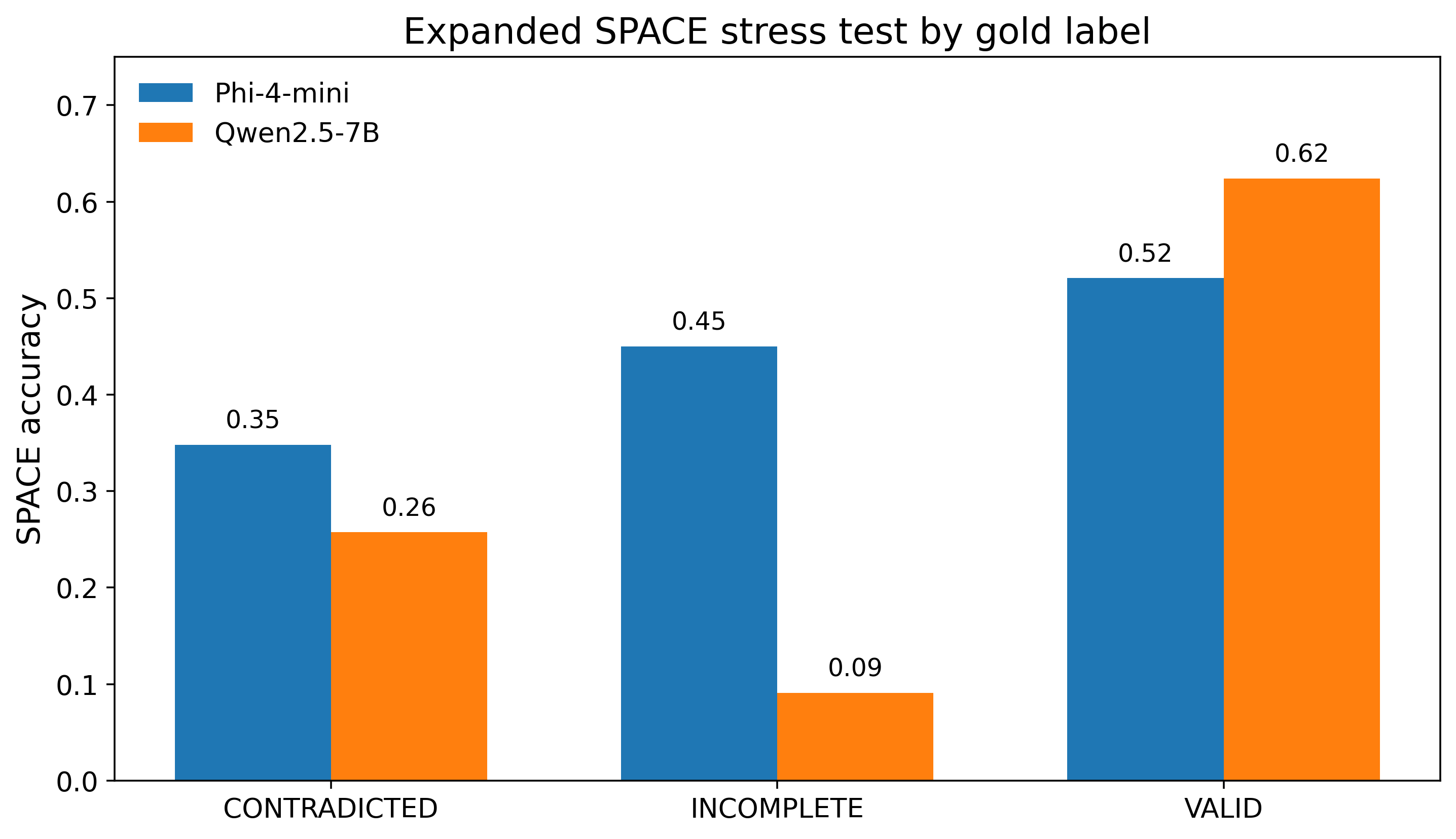}
    \caption{Expanded SPACE stress-test accuracy by gold SPACE label.}
    \label{fig:expanded_space_label}
\end{figure}

The primary benchmark evaluation treats original benchmark questions as \textit{VALID}. To evaluate whether SPACE labels can identify genuine answer-space defects, we use the expanded 12,000-example SPACE stress test described in the methodology. Table~\ref{tab:expanded_space_overall} reports the overall results, and Figure~\ref{fig:expanded_space_label} visualizes SPACE accuracy by gold label.

\begin{table}[t]
\centering
\caption{Expanded SPACE stress-test results across 12,000 controlled answer-space perturbations per model.}
\label{tab:expanded_space_overall}

\begin{tabular}{lrrr}
\toprule
Model & N & SPACE coverage & SPACE accuracy \\
\midrule
Qwen2.5-7B & 12000 & 1.0000 & 0.3074 \\
Phi-4-mini & 12000 & 0.9885 & 0.4168 \\
\bottomrule
\end{tabular}

\end{table}

Qwen2.5-7B achieves complete SPACE coverage but only 0.3074 overall SPACE accuracy. Phi-4-mini achieves 0.9885 SPACE coverage and 0.4168 SPACE accuracy. Qwen2.5-7B performs best on \textit{VALID} cases but performs poorly on \textit{INCOMPLETE} cases. Phi-4-mini is more balanced across labels but still reaches only moderate accuracy. Detailed label-level and stress-type tables are provided in the Appendix. These stress-test results support a cautious interpretation of SPACE: it provides observable structured reliability metadata, but prompt-only SPACE classification is not yet a solved capability.

\subsection{Structured Reliability Behavior}

\begin{figure}[t]
    \centering
    \includegraphics[width=\linewidth]{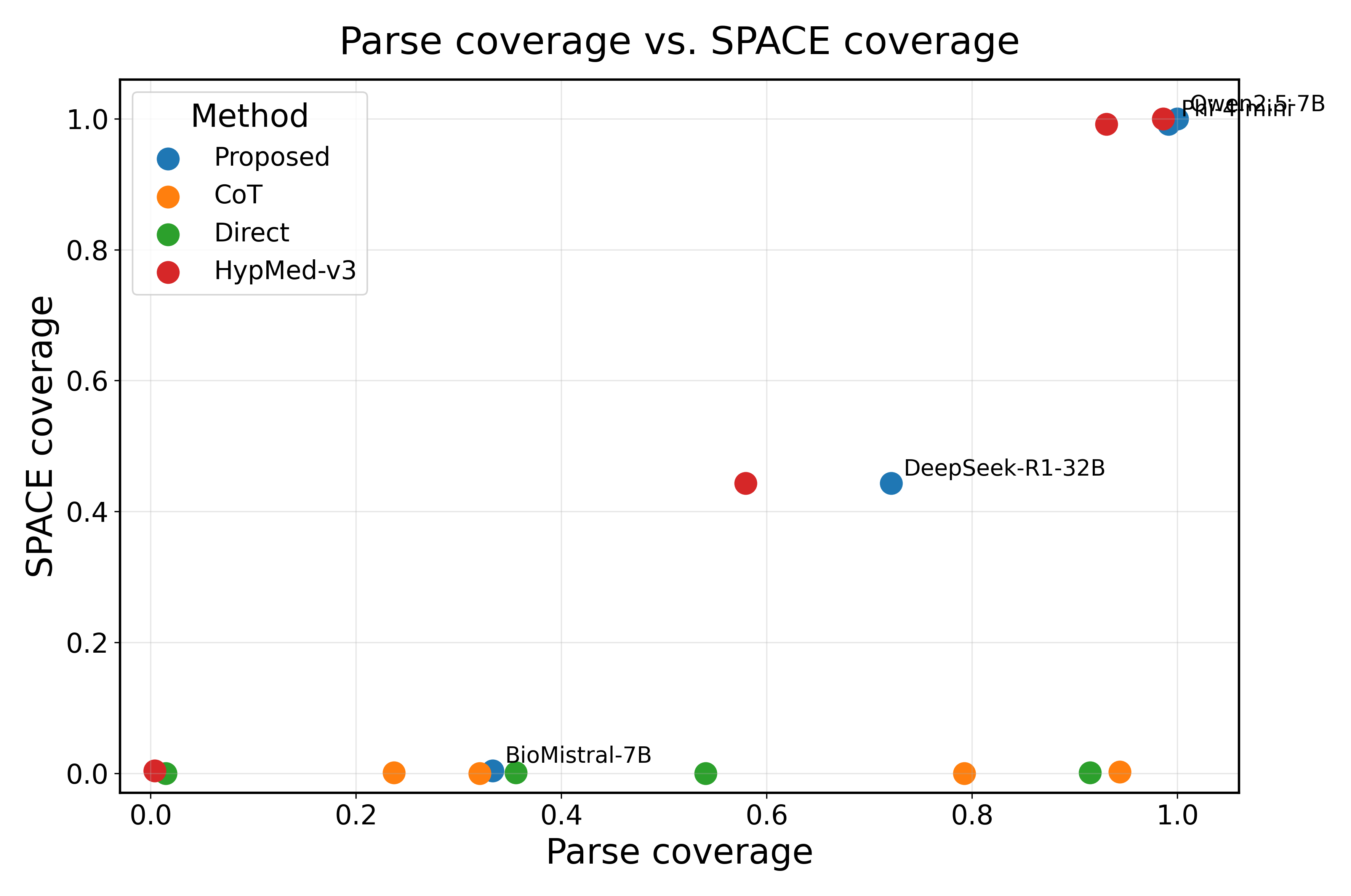}
    \caption{Parse coverage versus SPACE coverage across evaluated model-method combinations.}
    \label{fig:coverage}
\end{figure}

Figure~\ref{fig:coverage} compares parse coverage and SPACE coverage across evaluated model-method combinations in the primary experiment. The proposed pipeline produces high parse and SPACE coverage for Qwen2.5-7B and Phi-4-mini. DeepSeek-R1-32B and BioMistral-7B show weaker structured-output compliance, indicating that biomedical answer selection and machine-parseable reliability reporting are distinct capabilities.

False-commitment behavior further supports this separation. In the primary four-model experiment, the proposed method substantially reduces false commitment compared with direct and chain-of-thought prompting for Qwen2.5-7B and Phi-4-mini. The scaled results show the same general pattern: fusion reduces high-confidence wrong-answer behavior compared with direct and chain-of-thought prompting because reliability metadata is supplied by HypothesisMed-v3. Additional structured-output failure diagnostics are provided in the Appendix.

\subsection{Confidence and Calibration Analysis}

We evaluate calibration on the scaled fusion outputs using ten-bin expected calibration error. Table~\ref{tab:scaled_calibration} reports the scaled calibration summary. The scaled calibration results confirm that the confidence field is not well calibrated. For Qwen2.5-7B, ECE-10 ranges from 0.5306 to 0.6566 across datasets. For Phi-4-mini, ECE-10 ranges from 0.4311 to 0.6333. These values support using false commitment as a thresholded behavioral diagnostic rather than treating the confidence score as a calibrated probability.

\begin{table*}[t]
\centering
\caption{Scaled fusion calibration summary. ECE-10 denotes expected calibration error using ten confidence bins. For Phi-4-mini, $N$ is slightly below the dataset size in some rows because calibration is computed only over parseable fusion outputs.}
\label{tab:scaled_calibration}

\begin{tabular}{llrrrrr}
\toprule
Model & Dataset & N & Accuracy & Mean confidence & ECE-10 & Brier \\
\midrule
Qwen2.5-7B & MedQA & 5000 & 0.5930 & 0.1577 & 0.5375 & 0.5382 \\
Qwen2.5-7B & MedMCQA & 4183 & 0.5604 & 0.0766 & 0.5306 & 0.5306 \\
Qwen2.5-7B & PubMedQA & 1000 & 0.7530 & 0.2396 & 0.6566 & 0.6571 \\
Phi-4-mini & MedQA & 4937 & 0.5169 & 0.3261 & 0.4311 & 0.4475 \\
Phi-4-mini & MedMCQA & 4174 & 0.4854 & 0.3534 & 0.4524 & 0.4618 \\
Phi-4-mini & PubMedQA & 999 & 0.7097 & 0.0983 & 0.6333 & 0.6355 \\
\bottomrule
\end{tabular}

\end{table*}

\subsection{Positioning Against Published Results}

A conservative literature-positioning analysis shows that the proposed Qwen2.5-7B pipeline is competitive with reported Qwen2.5-7B biomedical QA results under prior published settings, although direct comparison is limited by differences in splits, prompts, shots, and decoding procedures \cite{yang2024qwen25}. The proposed method is above the reported Qwen2.5-7B chain-of-thought value on MedQA and MedMCQA and slightly below the reported PubMedQA value in the primary 1,000-example evaluation. Therefore, the supported claim is not universal state-of-the-art performance. The supported claim is that the proposed inference-time reliability pipeline is competitive with reported Qwen2.5-7B biomedical QA performance while adding structured SPACE reporting, parseability measurement, false-commitment analysis, scaled evaluation, and expanded answer-space stress testing.

\section{Discussion, Limitations, and Future Work}
\label{sec:discussion}

The results show that answer accuracy, parseability, structured SPACE reporting, confidence behavior, and false commitment are separable capabilities in biomedical QA. In the primary four-model experiment, the proposed answer-fusion plus HypothesisMed-v3 SPACE pipeline improves weighted answer accuracy over each model's best direct or chain-of-thought baseline while also increasing machine-parseable reliability reporting. The scaled evaluation provides a more nuanced conclusion. For Phi-4-mini, fusion substantially improves weighted accuracy over the best direct or chain-of-thought baseline, from 0.4296 to 0.5192 across 10,183 examples. For Qwen2.5-7B, chain-of-thought prompting remains slightly higher in scaled answer accuracy, with 0.6079 compared with 0.5953 for fusion. Nevertheless, the Qwen2.5-7B fusion pipeline provides complete parse coverage, complete SPACE coverage, and a much lower false-commitment rate than the direct and chain-of-thought baselines. Thus, the proposed pipeline should not be interpreted only as an accuracy-improvement method. Its broader contribution is to make biomedical QA outputs more observable, parseable, and auditable.

These findings extend prior inference-time reasoning work by showing that answer aggregation and reasoning prompts are not sufficient by themselves for reliability-sensitive biomedical workflows. Chain-of-thought prompting and self-consistency-style inference can improve answer selection, but they do not directly measure whether an output is machine-parseable, whether the answer space appears valid, or whether an incorrect answer is delivered with high confidence. The results also support broader multi-metric evaluation perspectives, which argue that accuracy should be evaluated alongside other reliability dimensions such as calibration, robustness, and transparency. In the biomedical setting, this distinction is important because a fluent or correct-looking response may still be difficult to audit or unsafe to use in an automated workflow.

The fusion ablation clarifies the role of each component. HypothesisMed-v3 alone is not consistently the strongest answer selector. Its main value is structured reliability reporting through SPACE labels and confidence values. Answer fusion contributes to answer selection by combining direct, chain-of-thought, and structured candidates, while HypothesisMed-v3 contributes answer-space metadata and false-commitment observability. This supports the design choice of separating answer selection from reliability reporting. The scaled fallback-order analysis further shows that the deterministic fallback order can affect accuracy, especially for Phi-4-mini. However, the larger pattern remains stable: fusion improves parseability and structured observability, while answer accuracy depends on model, dataset, and fallback behavior.

The expanded SPACE stress test addresses a key limitation of the original all-\textit{VALID} benchmark setting. The original benchmark evaluation measures whether models avoid falsely labeling standard questions as incomplete or contradicted, but it cannot fully test whether SPACE labels detect genuine answer-space defects. The expanded stress test introduces controlled \textit{INCOMPLETE} and \textit{CONTRADICTED} variants by removing the correct option or making the correct option non-unique. The results show that both Qwen2.5-7B and Phi-4-mini usually produce SPACE labels, but correct SPACE classification remains difficult. Qwen2.5-7B reaches 0.3074 SPACE accuracy on 12,000 stress-test examples, while Phi-4-mini reaches 0.4168. This finding shows that structured answer-space diagnosis is measurable, but it is not solved by prompt-only instruction following.

The calibration results further indicate that the confidence field should be interpreted cautiously. Scaled fusion outputs show high expected calibration error across datasets, meaning that the reported confidence values are not calibrated probabilities. Therefore, false commitment should be treated as a behavioral diagnostic for high-confidence wrong answers rather than as a complete uncertainty-calibration metric. The structured-output failure taxonomy also shows that failures arise from multiple sources, including answer parsing failures, missing SPACE labels, multiple JSON-like objects, and high-confidence wrong answers. These failure modes support the central argument that biomedical LLM reliability requires more than answer accuracy alone.

The results for DeepSeek-R1-32B and BioMistral-7B should also be interpreted cautiously. Their low parse coverage in several settings limits clean capability comparisons. However, these failures are still informative for workflow reliability. A model may have biomedical knowledge or reasoning capability but still be difficult to use in a strict machine-parseable reliability pipeline. This distinction is especially important for clinical and biomedical applications, where downstream systems often require structured outputs rather than free-form explanations alone.

This study has several limitations. First, the primary four-model benchmark evaluation uses 1,000 examples per dataset rather than full benchmark splits. We partially address this limitation through the scaled two-model evaluation, but the scaled evaluation is limited to Qwen2.5-7B and Phi-4-mini. Second, the expanded SPACE stress test is automatically constructed rather than expert annotated. Removing or duplicating answer options creates controlled defects, but real-world incomplete, ambiguous, contradicted, or evidence-conflict answer spaces may be more subtle. Third, confidence values are not calibrated probabilities, as shown by the scaled calibration analysis. Fourth, the study focuses on multiple-choice biomedical QA rather than open-ended clinical reasoning, clinical notes, retrieval-augmented generation, or deployment in real clinical workflows.

Future work should extend the evaluation to full benchmark splits, additional biomedical datasets, and more open-weight models. A key next step is to construct expert-validated SPACE examples with clinically realistic incomplete, ambiguous, contradicted, and evidence-conflict answer spaces. Future work should also evaluate constrained decoding, schema-guided generation, structured-output fine-tuning, and calibration methods. Finally, the same separation between answer generation, answer fusion, structured reliability reporting, and false-commitment analysis could be useful for biomedical assistant systems that require auditable outputs, including editorial-office and clinical-support question-answering workflows.

\section{Conclusion}
\label{sec:conclusion}
This paper presented HypothesisMed, an inference-time reliability pipeline for biomedical multiple-choice question answering. The method separates answer selection from reliability reporting by combining direct prompting, chain-of-thought prompting, HypothesisMed-v3 structured prompting, and answer fusion. In the primary four-model evaluation across MedQA, MedMCQA, and PubMedQA, the proposed answer-fusion plus HypothesisMed-v3 SPACE method improves weighted answer accuracy over each model's best direct or chain-of-thought baseline while adding structured SPACE reporting, parseability measurement, confidence metadata, and false-commitment analysis.

The scaled evaluation shows a more precise conclusion. For Phi-4-mini, fusion improves weighted accuracy substantially over the best direct or chain-of-thought baseline while also improving parse coverage and SPACE observability. For Qwen2.5-7B, chain-of-thought prompting remains slightly stronger in scaled answer accuracy, but fusion provides complete parse coverage, complete SPACE coverage, and substantially lower false commitment. Thus, HypothesisMed is best understood not simply as an accuracy-improvement method, but as a reliability and observability framework for biomedical QA workflows.

The expanded SPACE stress test further shows that SPACE-label prediction is not merely circular validation on all-\textit{VALID} benchmark items. By introducing controlled incomplete and contradicted answer spaces, the stress test demonstrates that current open-weight models can produce structured SPACE labels but still struggle to classify defective answer spaces reliably. This supports the broader finding that biomedical answer selection and structured answer-space diagnosis are distinct capabilities.

Overall, biomedical LLM reliability cannot be reduced to answer accuracy alone. Answer accuracy, parseability, SPACE-label coverage, SPACE-label accuracy, calibration behavior, and false commitment represent distinct evaluation axes. HypothesisMed provides a reproducible framework for evaluating biomedical LLMs as auditable workflow components rather than only as answer generators, and the results show both the promise and the current limitations of prompt-based structured reliability reporting.

\clearpage

\appendix
\section*{Appendix}
\addcontentsline{toc}{section}{Appendix}

\section{Code and Reproducibility Artifacts}

Code and reproducibility artifacts are available at 
\url{https://anonymous.4open.science/r/HypothesisMed-2B1E/}. The repository includes the source code, experiment scripts, final result tables, paper figures, model snapshot identifiers, inference settings, and compute-environment metadata. Raw benchmark datasets and raw model-output JSONL files are not included; users should obtain MedQA, MedMCQA, and PubMedQA from their official sources.

\section{HypothesisMed-v3 Prompt Template}
\label{app:prompt_template}

The HypothesisMed-v3 prompt asks the model to evaluate the answer space before selecting an answer and to return a machine-parseable structured response. The placeholders \texttt{\{question\}} and \texttt{\{options\}} are replaced with the benchmark question and answer choices.

\begin{quote}
\small
You are answering a biomedical multiple-choice question.

Evaluate whether the answer space is valid before selecting an answer.

\textbf{SPACE labels:}
\begin{itemize}
    \item \textbf{VALID}: the provided options contain one medically supported best answer.
    \item \textbf{INCOMPLETE}: the correct answer appears missing or insufficiently represented by the options.
    \item \textbf{CONTRADICTED}: the options are internally inconsistent, duplicated in a way that prevents a unique answer, or mutually contradictory.
\end{itemize}

Return only one JSON object with exactly these keys, where the confidence value is a number between 0.0 and 1.0:

\begin{verbatim}
{
  "space_label": "VALID|INCOMPLETE|CONTRADICTED",
  "answer": "A|B|C|D|E",
  "confidence": <number between 0.0 and 1.0>
}
\end{verbatim}

\textbf{Question:} \texttt{\{question\}}

\textbf{Options:} \texttt{\{options\}}
\end{quote}


\section{Scaled Evaluation and Extended Reliability Analyses}
\label{app:scaled_extended_results}

This appendix provides extended results supporting the scaled evaluation, paired significance testing, fallback-order sensitivity analysis, expanded SPACE stress testing, calibration analysis, and structured-output diagnostics. These results supplement the main findings and provide additional transparency about when the proposed pipeline improves answer accuracy, when it mainly improves observability, and where structured-output failures occur.

\begin{table}[H]
\centering
\caption{Weighted scaled aggregate results by model and method across MedQA, MedMCQA, and PubMedQA. The scaled evaluation uses 10,183 examples per model.}
\label{tab:app_scaled_weighted}
\resizebox{\textwidth}{!}{%
\begin{tabular}{llrrrrr}
\toprule
Model & Method & Accuracy & Parse cov. & SPACE cov. & SPACE acc. & False commit. \\
\midrule
Phi-4-mini & CoT & 0.4296 & 0.8062 & 0.0000 & 0.0000 & 0.9875 \\
Phi-4-mini & Direct & 0.2870 & 0.5567 & 0.0000 & 0.0000 & 0.8652 \\
Phi-4-mini & Fusion & 0.5192 & 0.9928 & 0.9905 & 0.4768 & 0.3271 \\
Phi-4-mini & HypMed-v3 & 0.3933 & 0.9074 & 0.9905 & 0.4768 & 0.3616 \\
Qwen2.5-7B & CoT & 0.6079 & 0.9967 & 0.0000 & 0.0000 & 0.9979 \\
Qwen2.5-7B & Direct & 0.5750 & 0.9917 & 0.0000 & 0.0000 & 0.9980 \\
Qwen2.5-7B & Fusion & 0.5953 & 1.0000 & 1.0000 & 0.5471 & 0.1146 \\
Qwen2.5-7B & HypMed-v3 & 0.4839 & 0.9785 & 1.0000 & 0.5471 & 0.0997 \\
\bottomrule
\end{tabular}
}

\vspace{1mm}
{\footnotesize
\textit{Notes.} Each model is evaluated on 10,183 examples across MedQA, MedMCQA, and PubMedQA. Parse cov. denotes parse coverage, SPACE cov. denotes SPACE-label coverage, SPACE acc. denotes SPACE-label accuracy, and false commit. denotes false commitment.
}
\end{table}

\begin{table}[H]
\centering
\caption{Scaled McNemar exact tests comparing the proposed fusion method with each model's chain-of-thought baseline.}
\label{tab:app_scaled_mcnemar}

\begin{tabular}{llrrrr}
\toprule
Model & Dataset & N & Fusion+/Base- & Base+/Fusion- & Exact $p$ \\
\midrule
Qwen2.5-7B & MedQA & 5000 & 273 & 383 & $2.00\times10^{-5}$ \\
Qwen2.5-7B & MedMCQA & 4183 & 265 & 278 & 0.607 \\
Qwen2.5-7B & PubMedQA & 1000 & 30 & 35 & 0.620 \\
Phi-4-mini & MedQA & 5000 & 578 & 127 & $1.48\times10^{-69}$ \\
Phi-4-mini & MedMCQA & 4183 & 607 & 142 & $3.03\times10^{-69}$ \\
Phi-4-mini & PubMedQA & 1000 & 27 & 31 & 0.694 \\
\bottomrule
\end{tabular}

\vspace{1mm}
{\footnotesize
\textit{Notes.} All comparisons use CoT as the baseline. Fusion+/Base- denotes examples where fusion is correct and the baseline is wrong. Base+/Fusion- denotes examples where the baseline is correct and fusion is wrong. CoT denotes chain-of-thought prompting.
}
\end{table}

\begin{xltabular}{\textwidth}{LLLrrr}
\caption{Scaled fallback-order sensitivity analysis. Each row reports the accuracy and parse coverage obtained when a different deterministic fallback order is used after majority voting fails to produce a strict majority.}
\label{tab:app_fallback_sensitivity}\\
\toprule
Model & Dataset & Fallback order & N & Parse cov. & Accuracy \\
\midrule
\endfirsthead

\caption[]{Scaled fallback-order sensitivity analysis. Continued.}\\
\toprule
Model & Dataset & Fallback order & N & Parse cov. & Accuracy \\
\midrule
\endhead

\midrule
\multicolumn{6}{r}{Continued on next page}\\
\endfoot

\bottomrule
\multicolumn{6}{p{\textwidth}}{\footnotesize \textit{Notes.} D = direct prompting; C = chain-of-thought prompting; H = HypothesisMed-v3 prompting.}\\
\endlastfoot

Qwen2.5-7B & MedQA & D $\rightarrow$ C $\rightarrow$ H & 5000 & 1.0000 & 0.5930 \\
Qwen2.5-7B & MedQA & D $\rightarrow$ H $\rightarrow$ C & 5000 & 1.0000 & 0.5894 \\
Qwen2.5-7B & MedQA & C $\rightarrow$ D $\rightarrow$ H & 5000 & 1.0000 & 0.6024 \\
Qwen2.5-7B & MedQA & C $\rightarrow$ H $\rightarrow$ D & 5000 & 1.0000 & 0.6014 \\
Qwen2.5-7B & MedQA & H $\rightarrow$ D $\rightarrow$ C & 5000 & 1.0000 & 0.5820 \\
Qwen2.5-7B & MedQA & H $\rightarrow$ C $\rightarrow$ D & 5000 & 1.0000 & 0.5820 \\
Qwen2.5-7B & MedMCQA & D $\rightarrow$ C $\rightarrow$ H & 4183 & 1.0000 & 0.5604 \\
Qwen2.5-7B & MedMCQA & D $\rightarrow$ H $\rightarrow$ C & 4183 & 1.0000 & 0.5580 \\
Qwen2.5-7B & MedMCQA & C $\rightarrow$ D $\rightarrow$ H & 4183 & 1.0000 & 0.5663 \\
Qwen2.5-7B & MedMCQA & C $\rightarrow$ H $\rightarrow$ D & 4183 & 1.0000 & 0.5663 \\
Qwen2.5-7B & MedMCQA & H $\rightarrow$ D $\rightarrow$ C & 4183 & 1.0000 & 0.5534 \\
Qwen2.5-7B & MedMCQA & H $\rightarrow$ C $\rightarrow$ D & 4183 & 1.0000 & 0.5551 \\
Qwen2.5-7B & PubMedQA & D $\rightarrow$ C $\rightarrow$ H & 1000 & 1.0000 & 0.7530 \\
Qwen2.5-7B & PubMedQA & D $\rightarrow$ H $\rightarrow$ C & 1000 & 1.0000 & 0.7530 \\
Qwen2.5-7B & PubMedQA & C $\rightarrow$ D $\rightarrow$ H & 1000 & 1.0000 & 0.7550 \\
Qwen2.5-7B & PubMedQA & C $\rightarrow$ H $\rightarrow$ D & 1000 & 1.0000 & 0.7550 \\
Qwen2.5-7B & PubMedQA & H $\rightarrow$ D $\rightarrow$ C & 1000 & 1.0000 & 0.7570 \\
Qwen2.5-7B & PubMedQA & H $\rightarrow$ C $\rightarrow$ D & 1000 & 1.0000 & 0.7570 \\
Phi-4-mini & MedQA & D $\rightarrow$ C $\rightarrow$ H & 5000 & 0.9874 & 0.5104 \\
Phi-4-mini & MedQA & D $\rightarrow$ H $\rightarrow$ C & 5000 & 0.9874 & 0.4772 \\
Phi-4-mini & MedQA & C $\rightarrow$ D $\rightarrow$ H & 5000 & 0.9874 & 0.5044 \\
Phi-4-mini & MedQA & C $\rightarrow$ H $\rightarrow$ D & 5000 & 0.9874 & 0.4962 \\
Phi-4-mini & MedQA & H $\rightarrow$ D $\rightarrow$ C & 5000 & 0.9874 & 0.4606 \\
Phi-4-mini & MedQA & H $\rightarrow$ C $\rightarrow$ D & 5000 & 0.9874 & 0.4586 \\
Phi-4-mini & MedMCQA & D $\rightarrow$ C $\rightarrow$ H & 4183 & 0.9978 & 0.4843 \\
Phi-4-mini & MedMCQA & D $\rightarrow$ H $\rightarrow$ C & 4183 & 0.9978 & 0.4674 \\
Phi-4-mini & MedMCQA & C $\rightarrow$ D $\rightarrow$ H & 4183 & 0.9978 & 0.4870 \\
Phi-4-mini & MedMCQA & C $\rightarrow$ H $\rightarrow$ D & 4183 & 0.9978 & 0.4757 \\
Phi-4-mini & MedMCQA & H $\rightarrow$ D $\rightarrow$ C & 4183 & 0.9978 & 0.4571 \\
Phi-4-mini & MedMCQA & H $\rightarrow$ C $\rightarrow$ D & 4183 & 0.9978 & 0.4580 \\
Phi-4-mini & PubMedQA & D $\rightarrow$ C $\rightarrow$ H & 1000 & 0.9990 & 0.7090 \\
Phi-4-mini & PubMedQA & D $\rightarrow$ H $\rightarrow$ C & 1000 & 0.9990 & 0.6450 \\
Phi-4-mini & PubMedQA & C $\rightarrow$ D $\rightarrow$ H & 1000 & 0.9990 & 0.7220 \\
Phi-4-mini & PubMedQA & C $\rightarrow$ H $\rightarrow$ D & 1000 & 0.9990 & 0.7250 \\
Phi-4-mini & PubMedQA & H $\rightarrow$ D $\rightarrow$ C & 1000 & 0.9990 & 0.6480 \\
Phi-4-mini & PubMedQA & H $\rightarrow$ C $\rightarrow$ D & 1000 & 0.9990 & 0.6570 \\
\end{xltabular}

\begin{table}[H]
\centering
\caption{Expanded SPACE stress-test results by gold SPACE label.}
\label{tab:app_space_stress_label}

\begin{tabular}{llrrr}
\toprule
Model & Gold SPACE & N & Coverage & Accuracy \\
\midrule
Qwen2.5-7B & CONTRADICTED & 6000 & 1.0000 & 0.2573 \\
Qwen2.5-7B & INCOMPLETE & 3000 & 1.0000 & 0.0910 \\
Qwen2.5-7B & VALID & 3000 & 1.0000 & 0.6240 \\
Phi-4-mini & CONTRADICTED & 6000 & 0.9837 & 0.3482 \\
Phi-4-mini & INCOMPLETE & 3000 & 0.9943 & 0.4500 \\
Phi-4-mini & VALID & 3000 & 0.9923 & 0.5207 \\
\bottomrule
\end{tabular}

\end{table}

\begin{table}[H]
\centering
\caption{Expanded SPACE stress-test results by controlled stress type.}
\label{tab:app_space_stress_type}

\begin{tabular}{llrrr}
\toprule
Model & Stress type & N & Coverage & Accuracy \\
\midrule
Qwen2.5-7B & Contradicted-Ambig. & 3000 & 1.0000 & 0.1320 \\
Qwen2.5-7B & Contradicted-Dup. & 3000 & 1.0000 & 0.3827 \\
Qwen2.5-7B & Incomplete-Remove & 3000 & 1.0000 & 0.0910 \\
Qwen2.5-7B & Valid-Original & 3000 & 1.0000 & 0.6240 \\
Phi-4-mini & Contradicted-Ambig. & 3000 & 0.9883 & 0.3560 \\
Phi-4-mini & Contradicted-Dup. & 3000 & 0.9790 & 0.3403 \\
Phi-4-mini & Incomplete-Remove & 3000 & 0.9943 & 0.4500 \\
Phi-4-mini & Valid-Original & 3000 & 0.9923 & 0.5207 \\
\bottomrule
\end{tabular}

\vspace{1mm}
{\footnotesize
\textit{Notes.} Contradicted-Ambig. denotes \texttt{CONTRADICTED\_AMBIGUOUS\_NONUNIQUE\_OPTION}; Contradicted-Dup. denotes \texttt{CONTRADICTED\_DUPLICATE\_CORRECT\_OPTION}; Incomplete-Remove denotes \texttt{INCOMPLETE\_REMOVE\_CORRECT}; Valid-Original denotes \texttt{VALID\_ORIGINAL}.
}
\end{table}

\begin{xltabular}{\textwidth}{LLLrrrrr}
\caption{Structured-output parsing failures across scaled evaluation outputs.}
\label{tab:app_parsing_failures}\\
\toprule
Model & Data & Method & N & AM & AMR & SM & SMR \\
\midrule
\endfirsthead

\caption[]{Structured-output parsing failures across scaled evaluation outputs. Continued.}\\
\toprule
Model & Data & Method & N & AM & AMR & SM & SMR \\
\midrule
\endhead

\midrule
\multicolumn{8}{r}{Continued on next page}\\
\endfoot

\bottomrule
\multicolumn{8}{p{\textwidth}}{\footnotesize \textit{Notes.} AM = answer missing; AMR = answer-missing rate; SM = SPACE missing; SMR = SPACE-missing rate. CoT denotes chain-of-thought prompting. HypMed-v3 denotes structured HypothesisMed-v3 prompting.}\\
\endlastfoot

Qwen2.5 & MedQA & Direct & 5000 & 51 & 0.0102 & 5000 & 1.0000 \\
Qwen2.5 & MedQA & CoT & 5000 & 26 & 0.0052 & 5000 & 1.0000 \\
Qwen2.5 & MedQA & HypMed-v3 & 5000 & 126 & 0.0252 & 0 & 0.0000 \\
Qwen2.5 & MedQA & Fusion & 5000 & 0 & 0.0000 & 0 & 0.0000 \\
Qwen2.5 & MedMCQA & Direct & 4183 & 34 & 0.0081 & 4183 & 1.0000 \\
Qwen2.5 & MedMCQA & CoT & 4183 & 7 & 0.0017 & 4183 & 1.0000 \\
Qwen2.5 & MedMCQA & HypMed-v3 & 4183 & 93 & 0.0222 & 0 & 0.0000 \\
Qwen2.5 & MedMCQA & Fusion & 4183 & 0 & 0.0000 & 0 & 0.0000 \\
Qwen2.5 & PubMedQA & Direct & 1000 & 0 & 0.0000 & 1000 & 1.0000 \\
Qwen2.5 & PubMedQA & CoT & 1000 & 0 & 0.0000 & 1000 & 1.0000 \\
Qwen2.5 & PubMedQA & HypMed-v3 & 1000 & 0 & 0.0000 & 0 & 0.0000 \\
Qwen2.5 & PubMedQA & Fusion & 1000 & 0 & 0.0000 & 0 & 0.0000 \\
Phi-4 & MedQA & Direct & 5000 & 2860 & 0.5720 & 5000 & 1.0000 \\
Phi-4 & MedQA & CoT & 5000 & 944 & 0.1888 & 5000 & 1.0000 \\
Phi-4 & MedQA & HypMed-v3 & 5000 & 492 & 0.0984 & 77 & 0.0154 \\
Phi-4 & MedQA & Fusion & 5000 & 63 & 0.0126 & 77 & 0.0154 \\
Phi-4 & MedMCQA & Direct & 4183 & 1423 & 0.3402 & 4183 & 1.0000 \\
Phi-4 & MedMCQA & CoT & 4183 & 1017 & 0.2431 & 4183 & 1.0000 \\
Phi-4 & MedMCQA & HypMed-v3 & 4183 & 130 & 0.0311 & 20 & 0.0048 \\
Phi-4 & MedMCQA & Fusion & 4183 & 9 & 0.0022 & 20 & 0.0048 \\
Phi-4 & PubMedQA & Direct & 1000 & 231 & 0.2310 & 1000 & 1.0000 \\
Phi-4 & PubMedQA & CoT & 1000 & 13 & 0.0130 & 1000 & 1.0000 \\
Phi-4 & PubMedQA & HypMed-v3 & 1000 & 321 & 0.3210 & 0 & 0.0000 \\
Phi-4 & PubMedQA & Fusion & 1000 & 1 & 0.0010 & 0 & 0.0000 \\

\end{xltabular}

\begin{xltabular}{\textwidth}{LLLrrrLr}
\caption{Answer correctness, multiple-JSON behavior, and high-confidence wrong outputs across scaled evaluation outputs.}
\label{tab:app_json_answer_behavior}\\
\toprule
Model & Data & Method & N & WPA & CPA & MJ/MJR & HCW \\
\midrule
\endfirsthead

\caption[]{Answer correctness, multiple-JSON behavior, and high-confidence wrong outputs across scaled evaluation outputs. Continued.}\\
\toprule
Model & Data & Method & N & WPA & CPA & MJ/MJR & HCW \\
\midrule
\endhead

\midrule
\multicolumn{8}{r}{Continued on next page}\\
\endfoot

\bottomrule
\multicolumn{8}{p{\textwidth}}{\footnotesize \textit{Notes.} WPA = wrong parsed answer; CPA = correct parsed answer; MJ/MJR = multiple JSON objects and multiple-JSON rate; HCW = high-confidence wrong. CoT denotes chain-of-thought prompting. HypMed-v3 denotes structured HypothesisMed-v3 prompting.}\\
\endlastfoot

Qwen2.5 & MedQA & Direct & 5000 & 2099 & 2850 & 16/0.0032 & 2095 \\
Qwen2.5 & MedQA & CoT & 5000 & 1899 & 3075 & 39/0.0078 & 1895 \\
Qwen2.5 & MedQA & HypMed-v3 & 5000 & 2576 & 2298 & 4980/0.9960 & 285 \\
Qwen2.5 & MedQA & Fusion & 5000 & 2035 & 2965 & 4980/0.9960 & 262 \\
Qwen2.5 & MedMCQA & Direct & 4183 & 1891 & 2258 & 11/0.0026 & 1886 \\
Qwen2.5 & MedMCQA & CoT & 4183 & 1819 & 2357 & 34/0.0081 & 1814 \\
Qwen2.5 & MedMCQA & HypMed-v3 & 4183 & 2155 & 1935 & 4179/0.9990 & 102 \\
Qwen2.5 & MedMCQA & Fusion & 4183 & 1839 & 2344 & 4179/0.9990 & 100 \\
Qwen2.5 & PubMedQA & Direct & 1000 & 253 & 747 & 0/0.0000 & 253 \\
Qwen2.5 & PubMedQA & CoT & 1000 & 242 & 758 & 0/0.0000 & 242 \\
Qwen2.5 & PubMedQA & HypMed-v3 & 1000 & 306 & 694 & 1000/1.0000 & 81 \\
Qwen2.5 & PubMedQA & Fusion & 1000 & 247 & 753 & 1000/1.0000 & 73 \\
Phi-4 & MedQA & Direct & 5000 & 1014 & 1126 & 2/0.0004 & 785 \\
Phi-4 & MedQA & CoT & 5000 & 1955 & 2101 & 0/0.0000 & 1939 \\
Phi-4 & MedQA & HypMed-v3 & 5000 & 2541 & 1967 & 4722/0.9444 & 1012 \\
Phi-4 & MedQA & Fusion & 5000 & 2385 & 2552 & 4722/0.9444 & 835 \\
Phi-4 & MedMCQA & Direct & 4183 & 1498 & 1262 & 2/0.0005 & 1414 \\
Phi-4 & MedMCQA & CoT & 4183 & 1605 & 1561 & 1/0.0002 & 1572 \\
Phi-4 & MedMCQA & HypMed-v3 & 4183 & 2285 & 1768 & 3390/0.8104 & 886 \\
Phi-4 & MedMCQA & Fusion & 4183 & 2148 & 2026 & 3390/0.8104 & 778 \\
Phi-4 & PubMedQA & Direct & 1000 & 234 & 535 & 0/0.0000 & 232 \\
Phi-4 & PubMedQA & CoT & 1000 & 274 & 713 & 0/0.0000 & 274 \\
Phi-4 & PubMedQA & HypMed-v3 & 1000 & 409 & 270 & 993/0.9930 & 28 \\
Phi-4 & PubMedQA & Fusion & 1000 & 290 & 709 & 993/0.9930 & 19 \\

\end{xltabular}

\end{document}